\documentclass[conference]{IEEEtran}
\IEEEoverridecommandlockouts
\usepackage{cite}
\usepackage{url}
\usepackage{amsmath,amssymb,amsfonts}
\usepackage{graphicx}
\usepackage{textcomp}
\usepackage{algorithm}
\usepackage{algpseudocode}
\usepackage{amsmath}
\usepackage{amssymb}
\usepackage{graphicx}
\usepackage{xcolor}
\def\BibTeX{{\rm B\kern-.05em{\sc i\kern-.025em b}\kern-.08em
    T\kern-.1667em\lower.7ex\hbox{E}\kern-.125emX}}
\begin{document}

\title{AutoGrid AI: Deep Reinforcement Learning Framework for Autonomous Microgrid Management\\
\thanks{This work was supported by RBC Borealis through the Let’s Solve it program.}
}

\makeatletter
\newcommand{\linebreakand}{%
  \end{@IEEEauthorhalign}
  \hfill\mbox{}\par
  \mbox{}\hfill\begin{@IEEEauthorhalign}
}

\makeatother

\author{

\IEEEauthorblockN{Kenny Guo\textsuperscript{*}}
\IEEEauthorblockA{\textit{Electrical and Computer Engineering} \\
\textit{University of Toronto}\\
Toronto, Canada \\
\texttt{\footnotesize kennyg.guo@mail.utoronto.ca}
}
\and
\IEEEauthorblockN{Nicholas Eckhert\textsuperscript{*}}
\IEEEauthorblockA{\textit{Industrial Engineering} \\
\textit{University of Toronto}\\
Toronto, Canada \\
\texttt{\footnotesize nick.eckhert@mail.utoronto.ca}
}
\and
\IEEEauthorblockN{Krish Chhajer\textsuperscript{*}}
\IEEEauthorblockA{\textit{Electrical and Computer Engineering} \\
\textit{University of Toronto}\\
Toronto, Canada \\
\texttt{\footnotesize krish.chhajer@mail.utoronto.ca}
}
\linebreakand
\IEEEauthorblockN{Luthira Abeykoon\textsuperscript{*}}
\IEEEauthorblockA{\textit{Electrical and Computer Engineering} \\
\textit{University of Toronto}\\
Toronto, Canada \\
\texttt{\footnotesize luthira.abeykoon@mail.utoronto.ca}
}

\and
\IEEEauthorblockN{Lorne Schell}
\IEEEauthorblockA{\textit{Staff Applied Research Scientist} \\
\textit{RBC Borealis}\\
Montreal, Canada \\
\texttt{\footnotesize lorne.schell@borealisai.com}
}

}
\maketitle
\begin{abstract}
We present a deep reinforcement learning-based framework for autonomous microgrid management. tailored for remote communities. Using deep reinforcement learning and time-series forecasting models, we optimize microgrid energy dispatch strategies to minimize costs and maximize the utilization of renewable energy sources such as solar and wind. Our approach integrates the transformer architecture for forecasting of renewable generation and a proximal-policy optimization (PPO) agent to make decisions in a simulated environment. Our experimental results demonstrate significant improvements in both energy efficiency and operational resilience when compared to traditional rule-based methods. This work contributes to advancing smart-grid technologies in pursuit of zero-carbon energy systems. We finally provide an open-source framework for simulating several microgrid environments.

\end{abstract}

% \begin{IEEEkeywords}
% component, formatting, style, styling, insert
% \end{IEEEkeywords}

\section{Introduction}
As part of the global response to the climate crisis, Canada, along with other countries around the world, aims to achieve net-zero carbon emission grid by 2050. However, a major hurdle to achieving this goal remains lowering the high dependence on non-renewable sources of energy such as fossil fuels and diesel. According to the International Energy Association [1], carbon emissions from global electricity generation remain the highest of any energy sector, contributing 13,800 million tonnes of carbon dioxide generation in 2024. Specifically, in Canada, about 75\% of remote communities are supplied electricity through diesel generators, and even more use fossil fuels for electric utilities [2]. This remains a major barrier in the country’s transition to renewable energy as these communities account for a large share of the country’s emissions. Microgrids offer a promising solution to limit reliance on fossil fuels.

\begin{figure}[htbp]
\centerline{
\includegraphics[width=0.45\textwidth]{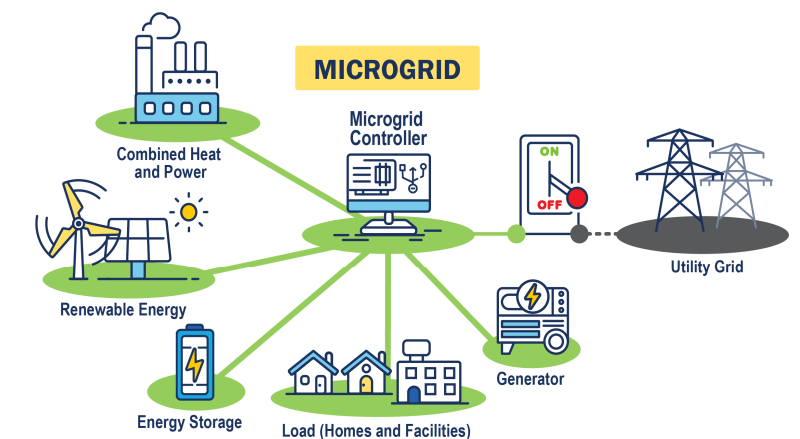}}
\caption{Different Components of a Microgrid}
\label{fig}
\end{figure}

Microgrids are local electricity grids, designed to supply electricity to a smaller-sized area, such as a residential neighborhood, town and industrial building(s). As shown in Figure 1 [3], a microgrid setup consists of several important electrical components, including distributed energy resources (DERs), energy storage systems (ESS), and residential or industrial loads. DERs refer to renewable energy generation sources, typically solar photovoltaic (PV) cells and wind turbines, along with non-renewable fuel generators, which together help meet local power demand. ESS refers to battery systems for storing excess energy or used as emergency supplies.

Microgrids can operate either parallel to the main grid, such as buying or selling local energy in case of abundance or shortage, or completely disconnecting from it by using only DERs to meet local demand (island mode). The microgrid strikes the perfect balance between meeting energy load demand and avoiding complete dependence on grid energy. A microgrid can offer increased energy efficiency, a reliable control system, and sustainable reduction in carbon emissions.

Despite their widespread adoption in Canada [4], rule-based microgrid algorithms face challenges in optimization due to highly fluctuating renewable energy generation. This uncertainty leads to constrained algorithms that must address supply/demand mismatch, energy waste, grid instability and overuse of ESS. Moreover, rule-based algorithms lack adaptability to real-time conditions [5][6][7]. Recent research suggests that reinforcement learning techniques may be utilized to produce better results than current rule-based techniques, due to their adaptability and potential for predicting complex relationships between a microgrid’s operating state and its environment variables. Deep reinforcement learning has consistently outperformed rule-based systems for microgrid management in recent studies. T. Nakabi [8] compared value-based and policy-based methods, showing A3C++ and PPO++ outperform traditional strategies. Other works explored DQN, SARSA, and Q-learning, highlighting DQN's strong performance [9]. 

Our goal is to:
\begin{enumerate}
    \item Build and train a reinforcement learning agent to manage microgrid systems, demonstrating robustness, adaptability, and advanced decision-making capabilities, validating past work. 
    \item Build upon past research by providing reinforcement learning agents with forecasting capabilities, leading to enhanced decision-making and grid robustness.
    \item Provide a framework to simulate microgrid environments using mathematical models and synthetic data generation techniques in cases of low data availability.
\end{enumerate}

\section{Data}
Data has been collected and processed from three different microgrids located in each the USA, Canada and Norway. This has been done to validate the agent's performance across varying conditions and environments.

\subsection{Mesa Del Sol Microgrid}

The Mesa Del Sol microgrid is located in Albuquerque, New Mexico, US, and designed to supply commercial and residential buildings [10]. Data from the Mesa Del Sol microgrid spans from July 2022 to July 2023 at a five-minute time interval. It includes measurements for several components including active power outputs of the battery, fuel cell generator, solar array, voltage frequency measurements, control signals for molded case circuit breakers (MCCB), the main grid interconnect power readings, and the individual power sources. 

To provide additional context, publicly available datasets were synthetically integrated, including the UC San Diego (UCSD) [11] energy load profile dataset, onshore wind farm generation data [12], and Ontario wholesale electricity prices from the Independent Electricity System Operator (IESO) [13]. All external services were truncated to the same 12-month window and aligned to a uniform five minute resolution via sliding window aggregation, linear interpolation, and preserving seasonal correlations across the different variables. Invalid variables, outliers, and irrelevant fields were discarded and removed. Energy load profiles were scaled such that the peak demand reached 80\% of the microgrid's maximum power generation capacity, in order to ensure feasible dispatch during RL simulations. 

% In parallel to this, five transformer-based forecasting models with a ten-step context window and a one-dimensional output were trained to predict solar generation, wind generation, critical load, non-critical load, and essential load, all of which achieved less than an MSE of 0.001. 

\subsection{Rye Microgrid}
The Rye Microgrid, a pilot within the EU’s research project REMOTE, is a small microgrid located in Langørgen, Norway. It is designed to supply energy to a modern farm and three households, to maximize the microgrid’s time in the island mode state. The environment consists of a 225kW wind turbine, an 86.4kWp PV system, and a battery ESS with a storage capacity of 500kWh. 
The dataset [14] consists of hourly sampled data from January 1st. 2020 to January 31st. 2021, with detailed solar generation, wind generation, total energy load, instantaneous grid price and sixteen weather parameters.
The dataset includes a testing set with hourly data from February 1st to March 8th, 2021. The test set exhibits significantly higher average load values compared to the training set. To ensure fair comparison, we scale the test set to match the maximum value distribution of the training dataset, eliminating bias introduced by the magnitude differences between datasets.

\subsection{Lac-Mégantic Microgrid}
 Synthetic data generation techniques were used to simulate the Lac-Mégantic Microgrid located in Quebec, chosen as the final environment. This microgrid serves a small town in rural Quebec and includes 30 buildings, 2,200 solar panels, and 700 kWh of battery storage. To generate synthetic data, we leveraged the Rye microgrid’s energy load and weather data for the town of Lac-Mégantic, sourced from OpenMeteo. Using this data, we generated load, energy production, and pricing information for this environment.

To create load data for Lac-Mégantic, the Rye microgrid load and weather data were used to train a random forest regressor to predict load for one building. After experimenting with several parameters, a simple model using the temperature, hour of the day, and day of the year was chosen. This model achieved an adjusted $R^2$ of 0.95 and a RMSE of 0.46. To obtain an estimated load for one building, we predicted a load using our model and applied a scaling factor of 1.5 to account for differences between lifestyle and energy usage in rural Norway versus Canada. To further simulate the differences between buildings, the prediction was scaled by a value sampled from a normal distribution with mean 1 ($\mu = 1$) and standard deviation 0.2 ($\sigma = 0.2$) and finally applied noise in the form of a value sampled from a normal distribution with mean 0 ($\mu = 0$) and standard deviation 0.1 ($\sigma = 0.1$). The process was then repeated 30 times per timestep to simulate the data for the 30 buildings within the Lac-Mégantic Microgrid.

Since Lac-Mégantic’s only sources of energy generation are various arrays of solar panels, mathematical modeling was used to predict energy generation. Satellite imaging was used to estimate the size of each solar panel and we obtained the energy efficiency from the Lac-Mégantic website. Using these calculations, solar irradiance measurements from OpenMeteo, and the following equation, energy generation was estimated.

\begin{equation}
    energy = \frac{irradiance * area * efficiency}{3600}
\end{equation}

Finally, hourly energy price were simulated using Hydro Quebec’s most common pricing plan, where the first 40 kilowatts purchased in a day are sold at \$0.06905 per kW and all kilowatts purchased after are sold at \$0.10652 per kW.

\section{Algorithm and Training}
% % SECTION # 2 - - - - - - - - - - - - - - - - - - - - - - - - - - - - 
To create the reinforcement learning agent, Proximal Policy Optimization (PPO) algorithm from Stable Baselines3's Python library (\texttt{sb3}) was employed \cite{b15}. PPO is an on-policy actor-critic method addressing instability issues of policy gradient methods by using a clipped surrogate objective function that constrains policy updates to prevent destructive large steps during training. PPO optimizes a combined objective function that incorporates policy improvement, function accuracy and exploration:
\begin{equation}
\small
L_{t}^{\text{CLIP+VF+S}}(\theta) = \hat{\mathbb{E}}_t\left[L_{t}^{\text{CLIP}}(\theta) - c_1 L_{t}^{\text{VF}}(\theta) + c_2 S[\pi_\theta](s_t)\right]
\end{equation}
The function consists of three main components: clipped policy loss, value function loss and entropy bonus.
The clipped policy loss prevents destructive policy updates by clipping probability between old and new policies.
\begin{equation}
\scalebox{0.85}{$\displaystyle
L^{\mathrm{CLIP}}(\theta)
= \hat{\mathbb{E}}_{t}\!\left[
    \min\Bigl(
      r_{t}(\theta)\,\hat{A}_{t},\;
      \mathrm{clip}\bigl(r_{t}(\theta),\,1-\epsilon,\,1+\epsilon\bigr)\,\hat{A}_{t}
    \Bigr)
\right]$}
\end{equation}
The value function loss is a standard mean squared error loss for training the value function:
\begin{equation}
L^{\text{VF}}(\theta) = \hat{\mathbb{E}}_t\left[(V_\theta(s_t) - V_t^{\text{target}})^2\right]
\end{equation}
The entropy bonus encourages exploration by penalizing overly deterministic policies:
\begin{equation}
\small
S[\pi_\theta](s) = -\mathbb{E}_{a \sim \pi_\theta}\left[\log \pi_\theta(a|s)\right]
\end{equation}
During training, PPO collects fixed-length rollout trajectories, computes advantages with generalized advantage estimation, then updates its policy and value networks via multiple epochs of mini-batch optimization with entropy regularization and adaptive KL divergence to ensure stable learning.
% - - - - - - - - - - - - - - - - - - - - - - - - - - - - - - - - - - - - - - - - - - 

\subsection{Mesa Del Sol Microgrid Environment}
The Mesa Del Sol environment replicates real-deployment scenarios. Training operates on weekly cycles: the agent observes 2016 timesteps (equivalent to 7 days at 5-minute intervals) before performing weight updates. This design ensures the agent learns from complete weekly patterns rather than shorter, potentially incomplete operational cycles.

In addition to the current timestep's critical load, solar, wind, critical, non-critical, and essential load data, five transformer-based forecasting models with a ten-step context window and a one-dimensional output were trained to predict solar generation, wind generation, critical load, non-critical load, and essential load. All models achieved an MSE less than 0.001. The PPO algorithm was implemented to include these forecasting models and initialize all five predictions for solar, wind, critical, non-critical, and essential load models for inference. The model takes in the current state of the microgrid from the dataset as part of its observation space. The model is also given context information about future forecasts to take more robust and efficient decisions. The model has four actions of continuous output, which are: the battery power (-1 to 1), fuel cell power (0 to 1), generator power (0 to 1) and island mode (0 to 1) decisions. The battery, fuel cell, and generator power action outputs allow the model to discharge/charge from the battery, at a fixed maximum rate, and also generate power at their respective maximum power rates. 

The model’s observed current state is analyzed by a reward function, calculated based on several different factors. The state of the microgrid is statically calculated based on predefined simulation rules, for example, the microgrid sells excess energy when connected to the grid. Some of the main factors contributing to positive reward include exporting excess energy to the main grid, the ratio of renewable to non-renewable energy generated/received, and the battery state of charge staying near half fully charged. Some negative rewards include not meeting load power demands, staying connected to the grid, and importing energy from the main grid. We exploited the reward function weights to get a reward for disconnecting from the grid, and also meeting load. 

Due to the need for custom functionality, a custom actor-critic policy network architecture was implemented to add dropout, which, at the time of this paper, is not currently part of the library's functionality. The policy network and value network process the 32-dimensional observation space through three fully connected layers with 512, 128, and 64 neurons, respectively. Each layer applies ReLU activation followed by dropout with a configurable dropout rate (typically 0.1-0.2). The value function network mirrors this architecture, ensuring equal policy optimization and value estimation.

The implementation also employs a linear learning rate scheduler that transitions from an initial rate to a final rate throughout training. The custom scheduler computes the current learning rate based on training progress. Custom callback functionality monitors policy standard deviation and enforces bounds to prevent exploding gradients. When the policy's standard deviation exceeds a maximum standard deviation, the callback resets it to a specific value by directly scaling down the weights of the network. This intervention maintains exploratory behavior throughout training, preventing early convergence to poor deterministic policies as well as exploding gradients. 

\subsection{Rye Microgrid Environment}
Similar to the reinforcement learning agent for Mesa Del Sol environment, proximal policy optimization has been employed from Stable Baselines 3’s Python library for the Rye environment. The agent learns in an OpenAI Gym environment, simulating a full year’s data of 2021, consisting of 8,760 hourly timesteps, taken from the Rye microgrid’s training set. The agent’s observation space consists of the current state of the microgrid solar energy production, wind energy production, load, instantaneous energy price, and current hour timestamp. In addition to these, the RL agent’s observation space also includes extra variables from transformer models, trained to forecast renewable energy production based on previous hourly generation readings. The agent’s action space consists of seven discrete actions, representing legal combinations of actions taken for the individual electrical components, i.e. renewable energy, battery and grid. This technique allows the agent to avoid taking undesirable actions, such as charging and discharging the battery at the same time.

% In action zero, renewable energy is used to charge the battery while grid is connected to satisfy load demands. In actions one and two, renewable energy is used to charge the battery and satisfy load at the same time, while the grid may be either connected or disconnected respectively. In actions three and four, both renewable energy and battery is used to satisy load demands while the grid is either connected or disconnected respectively. In both actions five and six the grid is connected to sell energy either by the renewable sources or the battery, whereas the other source is used to satisfy the load demands.

The environment’s reward function has been determined using a variety of factors: rewards for buying/selling from the grid, rewards for renewable energy usage and high penalties for not meeting user load demands. The rewards have been manually fine-tuned and scaled to ensure high efficiency and robustness of the model, trained for 100,000 timesteps.

\subsection{Lac-Mégantic Microgrid Environment}
Due to their similar natures, the Lac-Mégantic Environment utilizes a similar environment to Rye. Again, Stable Baselines 3's proximal policy optimization was used, as well as the same discrete seven action decision space and energy production prediction transformer. The Lac-Mégantic RL environment contains 8,784 steps accounting for a year of hourly data. The environment included variables such as solar production, energy consumption, battery state of charge, instantaneous energy price, and forecasted solar production. The reward for this environment is computed using a variety of factors: buying and selling rewards, total renewable energy rewards, unmet load penalties, and grid connection penalties. Tuning of these rewards resulted in a higher priority being assigned to grid connection penalties and unmet load penalties than other factors. After testing and tuning the reward function, the model was trained for 1,000,000+ timesteps.

\section{Results}
To provide a baseline to compare the models' performance, two baseline models were created: a simple rule-based algorithm and a more complex lookahead linear program. Both models operate on a simple action space, where the model can either charge or discharge the battery, buy or sell from the grid, or in the case of Mesa Del Sol, utilize the fuel cell or generator for additional energy generation. The rule-based algorithm makes a preset decision based on the circumstances of the timestep. This algorithm is both price and emission agnostic, but always makes sure that all load is met for the given timestep. If there is surplus energy and the battery is not fully charged, then it charges the battery. If there is surplus energy and the battery is fully charged, then the excess energy is sold to the grid. In cases of insufficient energy, first the battery is discharged and then energy is purchased from the grid. For Mesa Del Sol, the generator is first used to satisfy insufficient load, then the fuel cell, the battery, and finally the grid. This algorithm is feasible for use in the real world since it operates only on the current time steps and can make real time decisions. Our optimal baseline (lookahead linear program) operates by solving a mixed integer linear program (MILP) to optimize the cost of the microgrid for a week at a time. Unlike the rule-based algorithm, this program considers pricing but does not consider emissions. It is solved weekly using the PULP\_CBC\_CMD solver from the PULP library in Python with a 40-minute time limit per week, and serves purely as a baseline since it relies on future information and is not feasible for real-time use.

\subsection{Mesa Del Sol Microgrid}

\begin{figure}[htbp]
\centerline{
\includegraphics[width=0.45\textwidth]{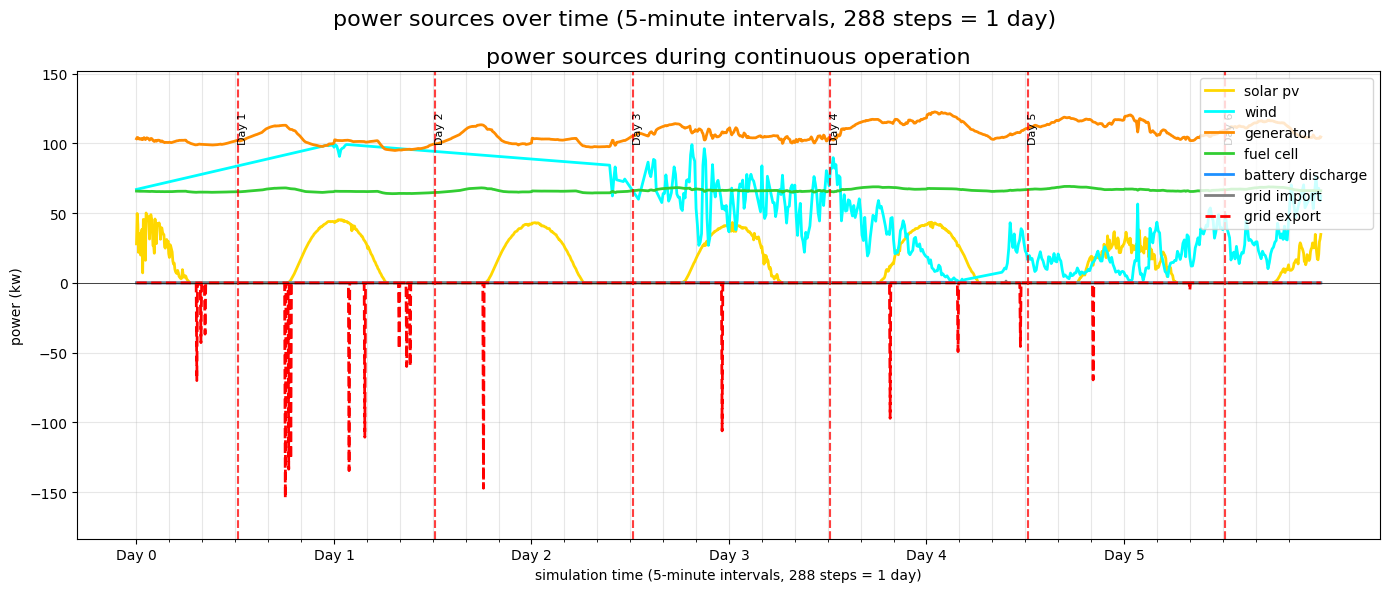}}
\caption{Mesa Del Sol State Space}
\label{fig}
\end{figure}

Looking at Figure 2, we can see that the trained agent learned a general pattern to increase fuel cell and generator power throughout the entire day. We believe that this pattern reflects the curve shape of the three loads, which tend to rise in the middle of the day as people are more likely to be actively using power during that time, and dropping during midnight hours. Additionally, we can also see the agent learned to charge the battery during peak hours. The scale of these changes are much smaller than expected and the model consistently outputs high power values, with small changes. This indicates the model highly prioritized disconnecting from the grid, and in order to be confident in its disconnection and avoid accumulating large negative penalties for not meeting load demands, the model decided to output high amounts of power in order to be confident. To potentially improve results, future testing could implement a new penalty for wasting power to guide the model to learn to closely match the total power demands at a given timestamp. Our agent and training methods improve on the current state of the art research, by utilizing forecasting models, to give more context to the model’s input. Our framework for this microgrid allows for constant testing and tuning through simulation, to allow for confident deployment in real-world settings.

\subsection{Rye Microgrid}

\begin{figure}[htbp]
\centerline{
\includegraphics[width=0.45\textwidth]{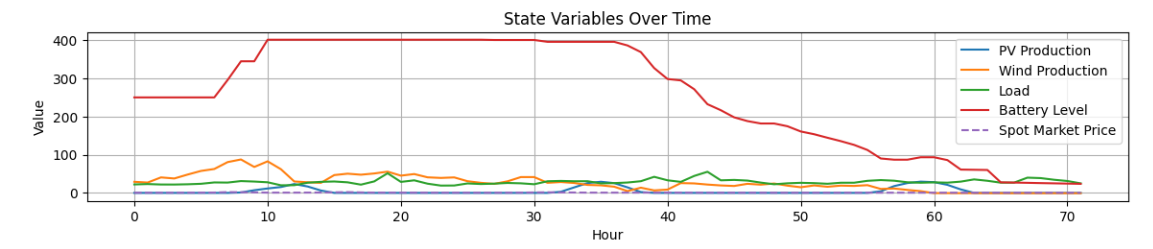}}
\centerline{
\includegraphics[width=0.45\textwidth]{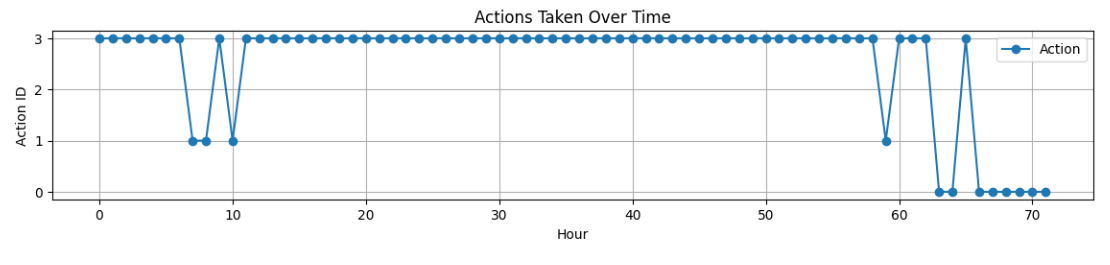}}
\caption{Rye Test Set State Space and Agent Actions}
\label{fig}
\end{figure}

As shown in Figure 3, the model learned a policy where it alternates between actions zero, one and three, charging the battery with excess surplus energy while using the battery or the connected grid to meet any excessive load demands. The microgrid also successfully meets 100\% of all load demands, doing so while maximising the time the microgrid is in island mode as favoured by the environment setting.

On the test set of the Rye dataset, the microgrid remains in island mode 71.43\% of the time, while meeting 30.61\% of total user demand with grid energy. While the cost of operating the microgrid is higher than the lookahead linear program and rule based program, using 32\% and 17.9\% of the grid energy respectively, this is due to the agent's policy of buying more from the grid rather than using the battery, in terms of excess load. However, the microgrid's island mode time is closer to the rule based program and higher than the lookahead linear program, both achieving 72.1\% and 35.1\% respectively, highlighting the model's robustness and bias towards meeting load demands while maximising islanded mode times.

\subsection{Lac-Mégantic}

\begin{figure}[htbp]
\centerline{
\includegraphics[width=0.45\textwidth]{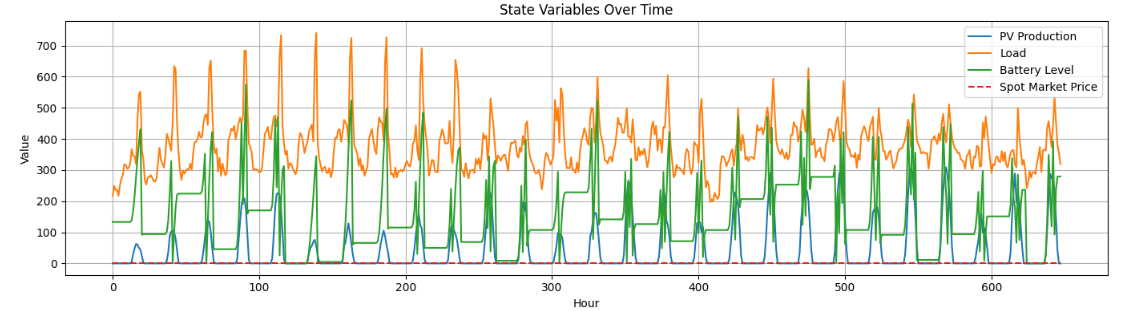}}
\centerline{
\includegraphics[width=0.45\textwidth]{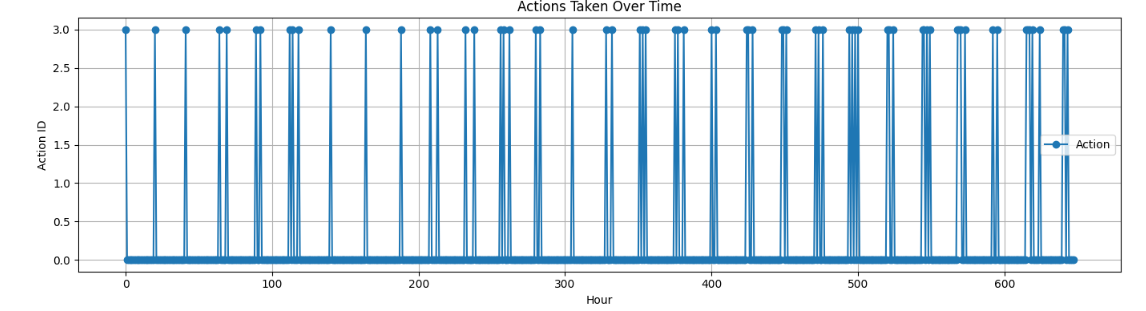}}
\caption{Lac-Mégantic State Space and Agent Actions}
\label{fig}
\end{figure}

After training, the model learned a policy where it alternates between actions zero and three. As seen in Figure 4, the model uses renewable energy to charge the battery during lower price and demand hours, while satisfying any load using the grid. Then, during peak hours the agent uses action three, where it satisfies the load using the saved up renewable energy. This action choice can be attributed to unique data distribution, where most timesteps feature a much higher consumption than generation, likely due to the 30 buildings with energy only being generated by solar panels. 

The final model achieved a lower cost and reduced reliance on the grid over the one-month test period compared to both baselines. The rule-based algorithm cost 24,024.44 CAD, satisfying 88.78\% of load with grid power and remaining in island mode 0.2\% of the time. The optimal baseline (PULP) cost 23,829.43 CAD, with 88.27\% grid load satisfaction and 0.7\% island mode. Our reinforcement learning (RL) model outperformed both on cost and island mode duration, costing 23,156.34 CAD, satisfying 88.69\% of load with the grid, and remaining in island mode 10.34\% of the time. Lower “load satisfied by the grid” means more demand met by renewable energy, assuming total load is fully met—unmet load would be undesirable. This shows the RL model operates more independently, making it more robust and cost-effective than the baselines.

\section{Conclusion}
We’ve demonstrated the practical viability of using RL optimization in realistic environments for microgrid mode optimization. This will drive overall effectiveness of microgrids across diverse installation configurations and help combat climate change.

\appendix
\renewcommand{\thefigure}{A\arabic{figure}}
\setcounter{figure}{0} % restart figure counter

\section*{Appendix A. Synthetic Data Generation Pipeline for Lac Mégantic Dataset}

The Figures ~\ref{fig:simload} and ~\ref{fig:solarpipe} below show the pipeline for energy consumption and solar energy synthetic data generation.

\begin{figure}[h!]
\centering
\includegraphics[width=\columnwidth]{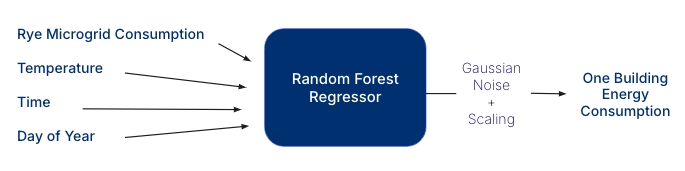}
\caption{Energy Consumption Generation Pipeline}
\label{fig:simload}

\end{figure}

\begin{figure}[h!]
\centering
\includegraphics[width=\columnwidth]{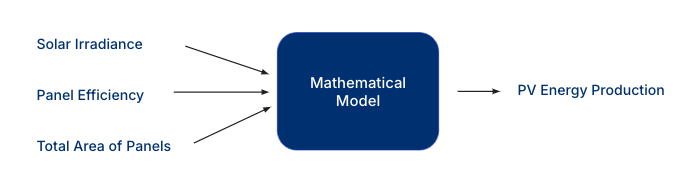}
\caption{Energy Consumption Generation Pipeline}
\label{fig:solarpipe}
\end{figure}

There is currently no public data available regarding energy consumption and production for Lac-Mégantic microgrid. While we were unable to numerically validate the correctness of synthetic data generated, the graph below in Figures ~\ref{fig:reallac} and ~\ref{fig:genlac}, taken from the microgrid's official website [16], serves as strong visual evidence of the generated data capturing the microgrid's seasonal consumption and generation trends.

\begin{figure}[h!]
\centering
\includegraphics[width=0.6\columnwidth]{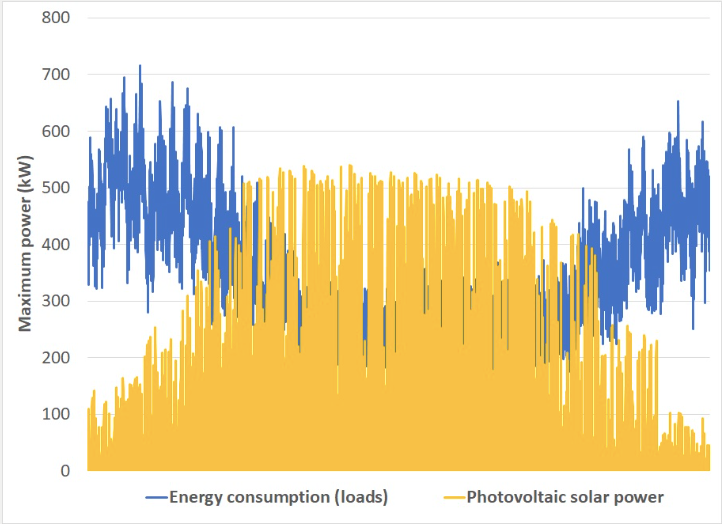}
\caption{Real Energy Production and Demand}
\label{fig:reallac}

\end{figure}

\begin{figure}[h!]
\centering
\includegraphics[width=0.6\columnwidth]{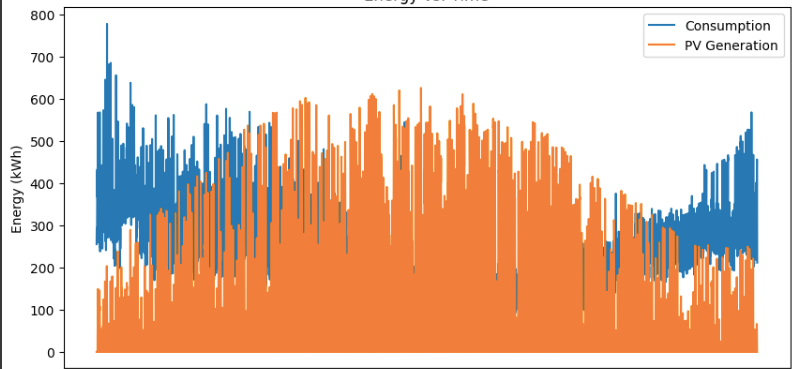}
\caption{Generated Energy Production and Demand}
\label{fig:genlac}
\end{figure}

\section*{Appendix B. Linear Program Formulation for Lookahead Linear Program}

The MILP maximizes a profit function while adhering to constraints for meeting load, maintaining a specific battery S.O.C. and logic constraints (i.e., cannot charge and discharge battery in the same step). Linearization techniques are used to make the program linear. Binary variables are used to represent each action during the $j$th timestep (where $x_{1j}$ is selling, $x_{2j}$ is buying, $x_{3j}$ is charging, $x_{4j}$ is discharging), continuous variables are used to represent the amount of energy used by each action during the $j$th time step ($y_{1j}$ is energy sold, $y_{2j}$ is energy bought, $y_{3j}$ is energy charged, and $y_{4j}$ is energy discharged) and corresponding activation variables ($q_{ij}$ $\forall$ $i \in \{1,2,3,4\}$). 

Additional variables represent other states of the system during the $j$th timestep including battery state of charge ($z_{1j}$), fuel cell discharge ($z_{2j}$) and generator discharge ($z_{3j}$) (generator and fuel cell variables are only included in the Mesa Del Sol formulation) with $C_{kj}$ indicating their maximum values. During the $j$th timestep, $P_j$ is the price of energy and $L_j$ is the load. Let $k \in \{1,2,3\}$, $i \in \{1,2,3,4\}$, $j \in \{1,2,\ldots,\text{number of timesteps}\}$. This program is formulated as follows, with $M$ indicating a sufficiently large number and $B$ indicating the battery capacity:

\begin{align*}
\text{Maximize} \quad & w = \sum_{j} [ P_j q_{1j} -  P_j q_{2j} + 0.1 z_{2j} + 0.1 z_{3j}] \\
\text{subject to} \quad 
& q_{1j}+q_{2j}+q_{3j}+q_{4j}+z_{2j}+z_{3j} = L_j \quad\forall j\\ 
& \sum_{i=1}^{4}x_{ij}=1 \quad\forall j \\
& z_{1,1}=0.5 \times B \\
& z_{1j} = z_{1,j-1} + q_{3j} - q_{4j}  \quad\forall j>1\\
& q_{ij} \leq y_{ij} \quad\forall i,j\\
& q_{ij} \geq y_{ij} - M(1-x_{ij}) \quad\forall i,j\\
& q_{ij} \leq M x_{ij} \quad\forall i,j\\
& 0 \leq z_{kj} \leq C_{kj} \quad\forall k,j\\
& y_{1,j},y_{2,j} \geq 0 \quad\forall j\\
& 0 \leq y_{3,j},y_{4,j} \leq B \quad\forall j\\
& x_{ij} \in \{0,1\} \quad\forall i,j\\
& q_{i,j} \geq 0 \quad\forall i,j
\end{align*}

This program is solved for each week using the PULP\_CBC\_CMD solver from the pulp library in Python. After each week is solved, the environment is updated with the optimal decisions and the window slides to the next week. Due to time and resource constraints, a 40 minute time limit was placed on the solver for each week. This program operates purely as a baseline since it can see future states and as a result is not feasible for real world use.

\end{document}